\documentclass{article}

\usepackage{bm}
\usepackage{amsmath}
\usepackage{amsthm}
\usepackage{wrapfig}
\usepackage{caption}
\usepackage{algorithm}
\usepackage{algorithmic}
\newtheorem{theorem}{Theorem}

\usepackage[nonatbib,preprint]{neurips_2022}

\usepackage[utf8]{inputenc} 
\usepackage[T1]{fontenc}    
\usepackage{hyperref}       
\usepackage{url}            
\usepackage{booktabs}       
\usepackage{amsfonts}       
\usepackage{nicefrac}       
\usepackage{microtype}      
\usepackage{xcolor}         
\usepackage{wrapfig}
\usepackage{color}
\usepackage{caption}
\usepackage{subfigure}
\usepackage{graphicx}
\usepackage{multirow}
\usepackage{wrapfig}
\usepackage{url}

\title{Counterfactual Neural Temporal Point Process for Estimating Causal Influence of Misinformation on Social Media}

\author{
  Yizhou Zhang\thanks{Equal Contribution} , Defu Cao\footnotemark[1] , Yan Liu\\
  Department of Computer Science\\
  Viterbi School of Engineering\\
  University of Southern California\\
  \texttt{\{zhangyiz,defucao,yanliu.cs\}@usc.edu} \\
}

\begin{document}

\maketitle

\begin{abstract}
  Recent years have witnessed the rise of misinformation campaigns that spread specific narratives on social media to manipulate public opinions on different areas, such as politics and healthcare. Consequently, an effective and efficient automatic methodology to estimate the influence of the misinformation on user beliefs and activities is needed. However, existing works on misinformation impact estimation either rely on small-scale psychological experiments or can only discover the correlation between user behaviour and misinformation. 
  To address these issues, in this paper, we build up a causal framework that model the causal effect of misinformation from the perspective of temporal point process. To adapt the large-scale data, we design an efficient yet precise way to estimate the \textbf{Individual Treatment Effect} (ITE) via neural temporal point process and gaussian mixture models. Extensive experiments on synthetic dataset verify the effectiveness and efficiency of our model. We further apply our model on a real-world dataset of social media posts and engagements about COVID-19 vaccines. The experimental results indicate that our model recognized identifiable causal effect of misinformation that hurts people's subjective emotions toward the vaccines.
\end{abstract}

\section{Introduction}
Recent researches reveals that widespread fake news and misleading information have been exploited by misinformation campaigns to manipulate public opinions in different areas, such as healthcare \cite{sharma2020coronavirus,sharma2021covid,kricorian2022covid} and politics \cite{luceri2020detecting}. To address this crucial challenge, research efforts from different perspectives have been devoted, such as fake news detection and coordination detection  \cite{sharma2019combating,sharma2020coronavirus,sharma2021covid}. 

However, an essential associated research question has not been explored sufficiently: how to know a piece of misinformation's causal influence on a user's beliefs and activities on a large-scale social media. Precisely estimating such impact is crucial for misinformation mitigation in various areas, e.g. delivering the corresponding clarification contents to the users that are most likely to be affected, allocating resources for more efficient and effective misinformation mitigation, and helping researchers understand misinformation campaigns better. Nevertheless, most of existing researches in social media analysis focus on understanding correlation between misinformation and user activities, rather than causal effect \cite{pearl2009causality, sharma2021covid,kricorian2022covid,wilson2020social}. As a result, they can not distinguish the effect from personal prior beliefs and engagement with misinformation. 
Current researches on misinformation's causal influence on people are mainly from psychology field \cite{jolley2014effects,van2015conspiracy}. They are usually based on carefully designed psychological randomised controlled trials on recruited subjects. Thus, it is impossible to extend them onto large-scale social media platforms due to the high cost to recruit enough subjects and ethical risk in conducting such a large-scale psychology experiments.

Since personal beliefs are usually unobservable, researchers usually apply the feature of the tweets generated or retweeted by the users as a proxy \cite{sharma2021covid}. However, the lack of appropriate algorithmic tools to conduct causal analysis on social media posts prevents researchers from understanding causal effect of misinformation. The processes that social media users generate original posts and engage with existing posts are typical temporal point processes. But existing methodologies for temporal causal effect estimation mostly focus on covariates and outcomes continuously distributed on timeline \cite{bica2020estimating,Bellot2021PolicyAU,bahadori2021debiasing, cao2020spectral,kamra2021polsird}, rather than discrete event points randomly scattered on timeline. Although the essential theory for counterfactual analysis of point process is already established\cite{roysland2012counterfactual}, most works are motivated by healthcare and thus focus on the hazard models, e.g. survival analysis \cite{aalen2020time} or the chance to catch cancer \cite{ryalen2020causal}, which only consider the single occurrence of the most recent future event. However, on social media, we care more about multiple events happening in a time window. \cite{gao2021causal} and \cite{noorbakhsh2021counterfactual} are rare works studying the causal effect on multiple occurrences in temporal point process. But \cite{noorbakhsh2021counterfactual} mainly focus on simulating the counterfactual events given an intervened intensity function rather than learning the treatment effect of specific factors on the process. As for \cite{gao2021causal}, one of its assumption is that the event marks must be categorized to a finite number of classes, leaving no space for the rich continuous features of social media posts, such as user sentiment and subjectivity scores.

In this work, we propose a framework that models the causal effect of a given piece of information on user beliefs and activities via counterfactual analysis on temporal point process \cite{sharma2020identifying,du2016recurrent,mei2017neural,zhangself,zuo2020transformer,omi2019fully,chen2020neural} with continuous features. We first define a causal structure model that characterizes the misinformation impact as how the engagement with the misinformation change a user's intensity function of generating original posts. In this model, the engagement with misinformation is considered as the treatment, and the user's future conditional intensity function is considered as the outcome. Then we design a functional that converts the change of two functions to a vector with intuitive physical meaning~\cite{meng2022physics}.  
To estimate the effect, we design a neural temporal point process model. It disentangles the distribution of event timestamp and post feature (e.g. the text embedding). Then it models the distribution of post features and event timestamp with Gaussian Mixture Model and temporal point process respectively. Such design enables it to acquire a closed-form solution of the feature expect without losing expressive power, leading to a balance between precision and efficiency. 

A critical challenge in training neural networks to recognize causal effect is the hidden bias in the dataset. In social media data, the most crucial bias is from information cocoons \cite{zuiderveen2016should}: users tend more to engage with the contents that they are interested in, and thus personalized recommendation systems will deliver user more contents that they are interested in to increase user engagement. 
Such bias leads to a data distribution different from randomised controlled trials and thus make neural networks give biased estimation. 
To decorrelate time-varying treatment from user's covariates and history in point process, 
we apply adversarial training to optimize a min-max game. More specifically, the encoder tries to minimize the likelihood of the observed treatments while a treatment predictor tries to maximize it. Our theoretic analysis proves that any balanced solution of the min-max game, rather than the global optimal solution in existing works \cite{bica2020estimating}, can help us remove the bias from information cocoons. In addition, the extensive experiments on synthetic datasets and real-world datasets indicate that our framework is able to 
approximate unbiased and identifiable estimation on the causal effect. In conclusion, the main contributions of the proposed model are as follows:

\begin{itemize}
    \item We propose a novel research problem on misinformation impact, which aims to find the causal effect of misinformation on users' belief and activities on social media.
    \item We propose a causal structure model to quantify the causal effect of misinformation and further design a neural temporal process model to conduct unbiased estimation to the effect.
    \item We evaluate our model on synthetic datasets to examine its effectiveness and efficiency and ues it to recognize identifiable causal effect of misinformation from real-world data.
\end{itemize}

\section{Related Work}

\subsection{Influence of Misinformation}
Recent researches about misinformation mainly focus on detecting fake news\cite{sharma2019combating,perez2017automatic,salem2019fa,sanaullah2022applications}, misinformation campaign detection \cite{sharma2020identifying,zhang2021vigdet} and understanding how fake news attract user engagement\cite{cheng2021causal,cheng2021causala}. Some researcher attempts to study the relation between misinformation and people's behaviours \cite{jolley2014effects,van2015conspiracy,sharma2021covid,kricorian2022covid}. However, most of them focus on mining the correlation between misinformation and people behaviours rather than causal effects. Only a limited amount of works, such as \cite{van2015conspiracy} try to understand the causal effect. However, they are usually from psychology field, and mainly rely on carefully designed randomised controlled trials. Extending such trials on large-scale social media platforms brings not only high cost but also potential ethical risk.
\subsection{Temporal Point Process}
The process that a user retweet or posts tweets can usually be modeled as a temporal point process with event feature \cite{sharma2020identifying, zuo2022differentially, samel2022learning,de2019temporal}. A temporal point process (TPP) with event feature is a stochastic process whose realization is a sequence of discrete events in a continuous timeline: $S = [(\bm{f}_1,t_1),(\bm{f}_2,t_2),...]$, where $\bm{f}$\footnote{We use bold font to emphasize that $\bm{f}$ is a feature vector rather than a function.} is the event feature (a scalar or a vector) and $t$ is the timestamp of the event. A TPP is fully characterized by an intensity function $\lambda(\bm{f},t|S_h)$ defined in the following integral equation:
\begin{equation}
    \mathbb{E}(N(\bm{F},T_1,T_2)|S_h)=\int_{\bm{F}}d\bm{f}\int_{T_1}^{T_2}\lambda(\bm{f},t|S_h)dt
    \label{eq:diff}
\end{equation}
where $\bm{F}$ is an area in the feature space, $S_h$ is the historical sequence of all events happening before time $T_1$, 
$N(\bm{F},T_1,T_2)$ is the number of events whose feature vectors are in $\bm{F}$ and timestamps are in the range $[T_1,T_2]$. The meaning of $\lambda(\bm{f},t|S_h)$ is the expected instantaneous speed that the user generate posts at point $\bm{f}$ in the feature space on time $t$. The process after time $t_i$ is fully described by $\lambda(\cdot,\cdot|S_h)$ \cite{chen2020neural}. 
Recent works propose to apply neural networks to model the $\lambda$ function \cite{du2016recurrent,mei2017neural,zhangself,zuo2020transformer,omi2019fully,chen2020neural}.

\subsection{Counterfactual Analysis on Temporal Point Process and Continuous Time Series}
The works focusing on studying the causal effect on multiple occurrences in temporal point process are rare. In \cite{noorbakhsh2021counterfactual}, the authors mainly focus on the sampling of counterfactual events rather than the learning the influence of specific factors on the intensity function. Another work\cite{gao2021causal} proposes a counterfactual analysis framework to understand the causal influence of event pairs in temporal point process. It defines the individual treatment effect (ITE) of an event toward future process as:
\begin{equation}
    ITE = \mu_y^1(t,t+T)-\mu_y^0(t,t+T) = \frac{1}{T}\int_t^{t+T}\lambda_y^1(t)-\lambda_y^0(t)dt
\end{equation}
where $\mu_y$ is the expect of the event number of type $y$ per unit time in time range $[t,t+T]$. $\mu_y^1$ indicate the case that a treatment is applied (exposed to misinformation) and $\mu_y^0$ is in contrary. However, this metric is only suitable in the case that the events can be categorized to finite discrete types. This is not applicable for social media post because most meaningful features of the posts, such as geographical information, sentiment score and subjective score, are naturally continuous. Forcibly discretizing them will lose meaningful information.  Besides, for counterfactual analysis of time series, ~\cite{bica2020estimating} proposes CRN, a neural model that can learn unbiased estimation to counterfactual world and causal effect. ~\cite{Bellot2021PolicyAU} proposes to analyze counterfactual estimation using synthetic controls via a novel neural controlled differential equation model. ~\cite{bahadori2021debiasing} introduces a new causal prior graph  to avoid the undesirable explanations that include confounding or noise and use a multivariate Gaussian distribution to model the real continuous values. However, all of them focus on modeling an observable variable existing on a continuous timeline instead of temporal point process. Unless getting heavily revised, such previous work can not be simply transferred to our problem scenario.

\section{Proposed Causal Structure Model and Treatment Effect}

\subsection{Causal Structure Model}
In this study, we focus on understanding how a user's engagement with the misinformation post causally influence the characters of the posts generated by the user in a fixed future time window. 
We formulate the process that a user interact the post shared by others and generate social media posts as two temporal point process where each event carries a continuous outcome vector. We denote the process of engaging the diffusion of a post as $P_{e}$ and the process generating new posts as $P_{g}$. Then the realization of the two temporal point process are respectively two sequences $S_{e}$ and $S_{g}$ of discrete events with continuous outcome vector in a continuous time range:
\begin{equation}
    S_{e} = [(\bm{f}^{(e)}_1,t^{(e)}_1),(\bm{f}^{(e)}_2,t^{(e)}_2),...],\quad S_{g} = [(\bm{f}^{(g)}_1,t^{(g)}_1),(\bm{f}^{(g)}_2,t^{(g)}_2),...]
\end{equation}

\begin{wrapfigure}{r}{0.62\textwidth}
\centering
    \includegraphics[width=0.6\textwidth]{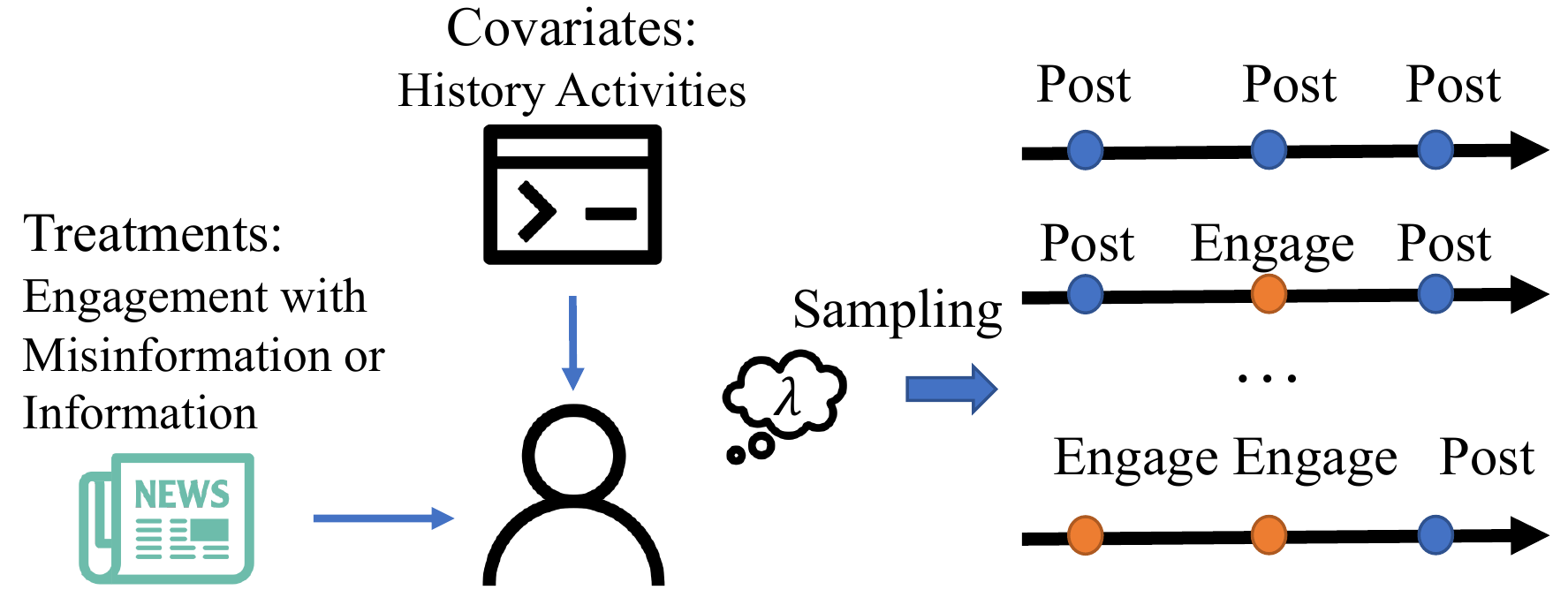}
    \caption{The proposed causal structured model describing the impact of a piece of information on user.}
    \label{fig:csm}
\end{wrapfigure}

where $\bm{f}$ is the feature vector characterizing the event and $t$ is the time stamp. For $S_{e}$, each event correspond to an interaction (e.g. "like`` or comment), and the vector $\bm{f}^{(e)}$ is the feature of the content (like the text representation, sentiment score and metadata) from others. Similarly, for $S_{g}$, the vector $\bm{f}^{(g)}$ is the feature of the content generated by the user. To examine the causal effect of an interaction event on the posts generated by the user in the future, we construct the following causal structure model, formulated as $<X,Y,Tr>$, where $X$ is the covariate, $Y$ is the outcome and $Tr$ is the treatment. In this model, given an interaction event $(\bm{f}^{(e)}_i,t^{(e)}_i)$ whose causal effect is to be examined, we consider the this event as the treatment $Tr$, and all the events, including both engagement events and posting events, that happen before $t^{(e)}_i$ are considered as the covariates. As for the outcome, rather than simply considering the most next generation event after $t^{(e)}_i$, we need a representation that can reflect the change of the whole generating process $S_g$ in a fixed future time window $T$. Thus we apply the conditional intensity function of the future process, denoted as $\lambda(\bm{f},t|Tr\cup X)$, as outcome. As discussed in the related work section, $\lambda$ function completely describe the future process. The overview of the model is presented in Figure. \ref{fig:csm}.

\subsection{Treatment Effect Evaluation}
In traditional counterfactual analysis works, the outcome is usually a scalar or a vector with finite dimensions. Thus, the treatment effect can be trivially computed by comparing the difference of the outcomes from real world and counterfactual world. However, in our framework, it is non-trivial to compute the difference of two functions. To overcome this challenge, we propose to first apply a functional $\mathcal{F}$ to project the $\lambda$ function to a vector with finite dimensions:
\begin{equation}
    \mathcal{F}_T(\lambda,Tr\cup X) = \frac{\phi(t,t+T,\lambda,Tr\cup X)}{\mu(t,t+T,\lambda,Tr\cup X)}
\end{equation}
\begin{equation}
    \phi(t,t+T,\lambda,Tr\cup X) = \mathbb{E}_{S\sim P(S|\lambda,Tr\cup X)}[\sum_{(\bm{f}_i,t_i)\in S_{t:t+T}}\bm{f}_i]
\end{equation}
\begin{equation}
\mu(t,t+T,\lambda,Tr\cup X) = \mathbb{E}_{S\sim P(S|\lambda,Tr\cup X)}|S_{t:t+T}|=\mathbb{E}(N(\text{sup}(\bm{f}),t,t+T)|Tr\cup X)
\end{equation}
where $P(S|\lambda,Tr\cup X)$ is the distribution of the event sequence $S$ sampled from the temporal point process described by $\lambda(\cdot,\cdot|Tr\cup X)$, $T$ is the time window that is a hyper-parameter, $\text{sup}(\bm{f})$ is the support set of $\bm{f}$ (the area where the probability density is larger than 0), and $S_{t:t+T}$ is a sub-sequence of $S$. $S_{t:t+T}$ contains every event in $S$ that happens at a time between $t$ and $t+T$. The intuitive meaning of $\mathcal{F}$ is the expected mean feature vector of all posts generated by a user. Thus, by comparing the outputs of $\mathcal{F}$ in real world and counterfactual world, we can see how the engagement with a specific post change the average features, e.g. general sentiment scores or text embedding. With this functional, we can simply compute the individual treatment effect as:
\begin{equation}
    ITE = \mathcal{F}_T(\lambda,Tr\cup X) - \mathcal{F}_T(\lambda,\emptyset\cup X)
\end{equation}
where $\emptyset$ is an empty set, $\mathcal{F}_T(\lambda,\emptyset\cup X)$ is the functional from counterfactual world in which we assume that the treatment is not applied (e.g. the misinformation post is not recommended or labeled as misinformation). For brief, we write $\mathcal{F}_T(\lambda,\emptyset\cup X)$ as $\mathcal{F}_T(\lambda,X)$ The overall impact of the treatment can be represented with the average treatment effect:
\begin{equation}
    ATE = \mathbb{E}_{(X,Tr)\sim U}[\mathcal{F}_T(\lambda,Tr\cup X) - \mathcal{F}_T(\lambda,X)]
\end{equation}
where $U$ is the set of users who engaged with the treatment post.

\subsection{Treatment Effect Calculation}
In the above sections, we define the causal structure model and the treatment effect. However, the above formulas are hard to compute. Therefore, in this subsection, we will derive a computable formulation of the treatment effect. We will start from the following theorem:
\begin{theorem}
For a user $u$, if the intensity function $\lambda(\bm{f},t|Tr\cup X)$ is known, then we have:
\begin{equation}
    \mu(t,t+T,\lambda_1,Tr\cup X) = \int_{\text{sup}(\bm{f})}d\bm{f}\int_t^{t+T}\lambda(\bm{f},t|Tr\cup X)dt
\end{equation}
\begin{equation}
    \phi(t,t+T,\lambda_1,Tr\cup X) = \int_{\text{sup}(\bm{f})}\bm{f}d\bm{f}\int_t^{t+T}\lambda(\bm{f},t|Tr\cup X)dt
\end{equation}
\end{theorem}
The first equation can be trivially proved by replacing the $\bm{F}$ in Equation \ref{eq:diff} with the support set. The second one can be proved with the Campbell's Theorem \cite{campbell:01}. A detailed proof is provided in the Appendix \ref{theo:1}.
The above formulas contain double integral, which is inefficient to compute. To transform the double integral to a single integral, based on a previous work in spatial-temporal point process \cite{chen2020neural}, we have:
\begin{equation}
    \lambda(\bm{f},t|Tr\cup X) = \lambda(t|Tr\cup X)p(\bm{f}|t,Tr\cup X)
\end{equation}
Thus, we can disentangle $\lambda(\bm{f},t)$ and respectively model $\lambda(t_i)$ and $p(\bm{f}|t)$. More importantly, we can simply model $\mu$ as:
\begin{equation}
    \mu(t,t+T,\lambda,Tr\cup X) =\int_t^{t+T}\lambda(t|Tr\cup X)dt\int_{sup(\bm{f})}p(\bm{f}|t,Tr\cup X)d\bm{f}= \int_t^{t+T}\lambda(t|Tr\cup X)dt
\end{equation}
And with this formula, we have:
\begin{equation}
\begin{aligned}
    \phi(t,t+T,\lambda,Tr\cup X) 
    &=\int_{sup(\bm{f})}\int_t^{t+T}\bm{f}\lambda(t|Tr\cup X)p(\bm{f}|t,Tr\cup X)d\bm{f}dt\\
    &=\int_t^{t+T}\lambda(t|Tr\cup X)dt\int_{sup(\bm{f})}\bm{f}p(\bm{f}|t,Tr\cup X)d\bm{f} \\
    &= \int_t^{t+T}\lambda(t|Tr\cup X)\mathbb{E}[\bm{f}|t,Tr\cup X]dt
\end{aligned}
\end{equation}
The above formulas contain only single integrals. Thus, they can be efficiently approximated with summation: $\int_{x_1}^{x_2} f(x)dx\approx \sum_{i=0}^{(x_2-x_1)/\Delta x}f(x_1+i\Delta x)\Delta x$

\section{Neural Estimation of Treatment Effects}

The above section construct a causal framework that can measure the impact of a given social media post based on the change of $\lambda(t|Tr\cup X)$ and $p(\bm{f}|t,Tr\cup X)$. In this section, as shown in Figure ~\ref{fig:model} we will further discuss how to estimate the impact with a neural temporal point process model.

\subsection{Learning Conditional Intensity Function via Maximum Likelihood Estimation}
The log-likelihood of an observed event $(f,t)$ (no matter an engagement event or an generation event) can be written as:
\begin{equation}
    \log p(\bm{f},t|Tr\cup X)=\log\lambda(t|Tr\cup X)-\int_{t_n}^{t}\lambda(t|Tr\cup X)dt +\log p(\bm{f}|t,Tr\cup X)
\end{equation}
where $t_n$ is the timestamp of the last event in the set $Tr\cup X$. The above equation provides us with a way to learn $\lambda(t|Tr\cup X)$ and $p(\bm{f}|t,Tr\cup X)$ by maximizing the likelihood of each event given the historical information (treatment and covariates). To enable the model to make correct prediction for both $\lambda(\bm{f},t|Tr\cup X)$ and $\lambda(\bm{f},t|X)$, we construct two kinds of samples to train the functions:

\begin{wrapfigure}{r}{0.62\textwidth}
\centering
    \includegraphics[width=0.6\textwidth]{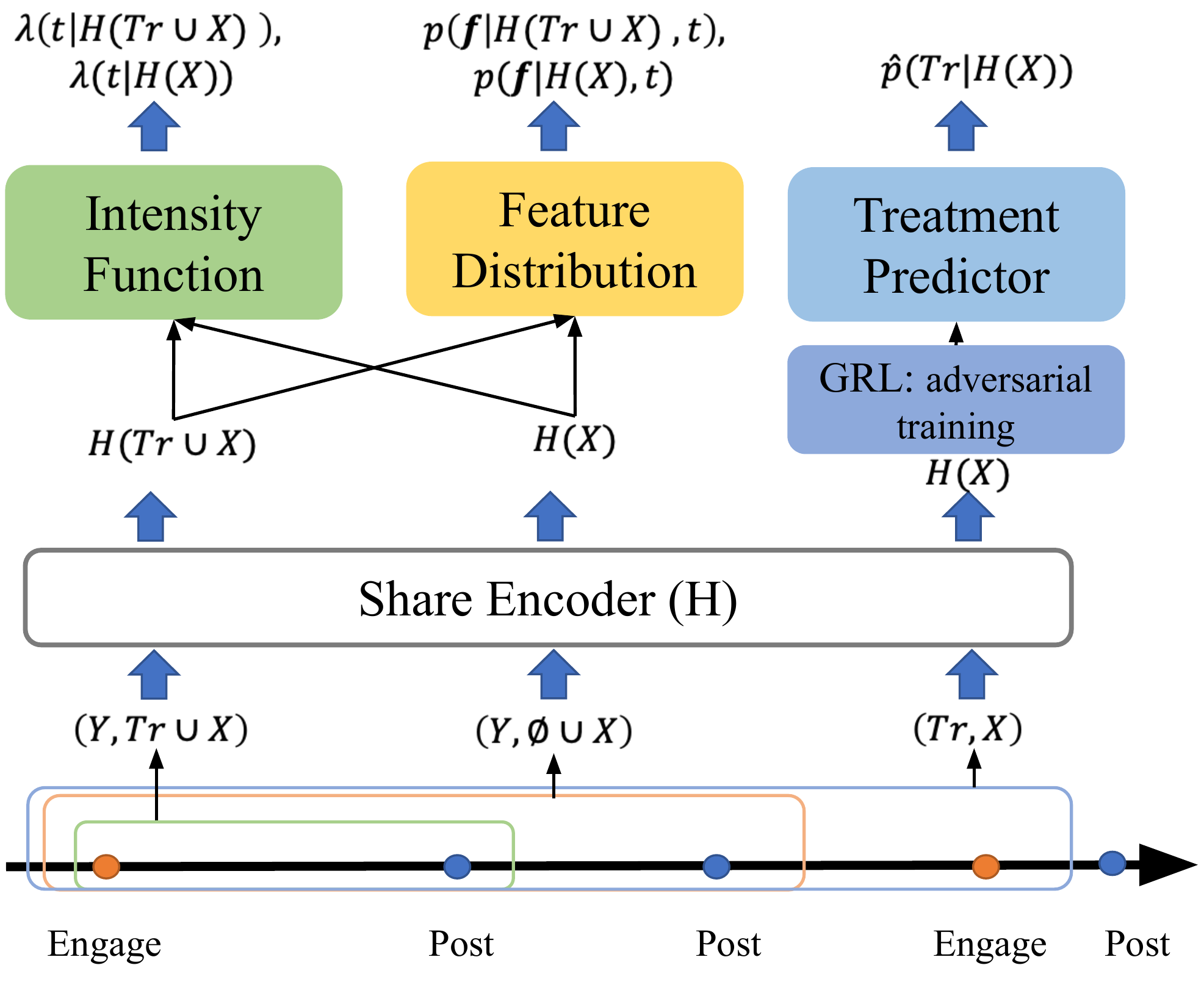}
    \caption{The proposed neural model to estimate the impact of misinformation.}
    \label{fig:model}
\end{wrapfigure}

\textbf{Samples with valid Treatment}: If for a generating event $(\bm{f}^{(g)},t^{(g)})$, its most recent previous event is an engagement event $(\bm{f}^{(e)},t^{(e)})$ (in other words, the user does not have other activity between the engagement event and the generating event), then we can construct a sample $(Y,Tr\cup X)$, where $Y=(\bm{f}^{(g)},t^{(g)})$, $Tr=(\bm{f}^{(e)},t^{(e)})$, and $X$ is a sequence that contains all engagement events and generation events before $Tr$.

\textbf{Samples without Treatment}: If for a generating event $(\bm{f}^{(g)},t^{(g)})$, its most recent previous event is still a generating event (in other words, the user generate two original posts without engaging with other posts), then we can construct a sample $(Y, X)$, where $Y=(\bm{f}^{(g)},t^{(g)})$ and $X$ is a sequence that contains all engagement events and generation events before $Y$. In this sample, between the last generation event in $X$ and $Y$, there is no interruption from treatment. Thus, it helps the model to learn $\lambda(t|X)$ and $p(\bm{f}|t,X)$

For a sample $(Y,Tr\cup X)$ or $(Y,X)$, we first use a shared encoder $H(\cdot)$ to project $(Tr\cup X)$ (or $X$) to a representation vector $h=H(Tr\cup X)$ (or $h = H(X)$ for the case where treatment is NULL). Then we model the intensity function and feature distribution as $\lambda(t|h)$ and $p(\bm{f}|t,h)$. For $\lambda(t|h)$, following FullyNN, we use a multi-layer perceptron $MLP(h,t)$ to model its integral $\int_{t_n}^{t}\lambda(t|h)dt$. The MLP's partial derivative with respect to $t$ is $\lambda(t|h)$. 

To model $p(\bm{f}|t,h)$, a straightforward solution borrowed from generative deep learning is to apply a neural network, i.e. the decoder, to transform a simple distribution, e.g. a Gaussian distribution whose parameters are decided based on $h$ and $t$, to a complicated distribution. The decoder can be trained via different loss function, like reconstruction error (variational auto encoder) and likelihood (normalizing flow)\footnote{GAN is not suitable because for fixed $h$ and $t$ we only have one sample, which can be easily memorized.} \cite{wang2020csan,chen2020neural}. However, this method has an important drawback: its conditional expect $\mathbb{E}[\bm{f}|t,h]$ does not have a closed-form solution. To compute the expect, we can only apply sampling or approximation, e.g. forwarding the expect of the simple distribution into the decoder\footnote{Because the decoder $D$ is a non-linear function, $\mathbb{E}[D(x)]$ is usually different from $D(\mathbb{E}[x])$}. 
To address the above challenge, we propose to explicitly model $p(\bm{f}|t,h)$ with a mixture of Gaussian distributions:
\begin{equation}
    p(\bm{f}|t,h) = \sum_{j=1}^mw_j(t,h)g(\frac{\bm{f}_i-\bm{u}_j(t,h)}{\sigma_j(t,h)})
\end{equation}
\begin{equation}
    w(t,h) = \text{softmax}(MLP_w(h,t)),\sigma(t,h) = \exp(MLP_\sigma(h,t)),\bm{\mu}_j(t,h)=MLP^{(j)}_{\bm{\mu}}(h,t)
\end{equation}
where $w_j$ is the mixture weight, $\sigma_j$ is a scalar, $\bm{u}_j$ is a vector with the same dimension as $\bm{f}$, and $g(\cdot)$ is a standard multivariate gaussian distribution $\mathcal{N}(\bm{0},\bm{I})$ whose covariant matrix is an identical matrix. Although the formula of each component is simple, their mixture has a theoretical guarantee on universal approximation to all distributions \cite{ZEEVI199799}. Because the expect of a mixture distribution is the mixture of the expects, we have a closed-form solution for $\mathbb{E}[\bm{f}|t,h]$:
\begin{equation}
    \mathbb{E}_{\bm{f}\sim p(\bm{f}|t,h)}[\bm{f}|t,h] = \sum_{j=1}^mw_j(t,h)\bm{u}_j(t,h)
\end{equation}

\subsection{Adversarial Balanced Neural Temporal Point Process}

As discussed in the related work section, by maximizing the likelihood of the posts generated by the users, we can train a neural network that predict $\lambda(t|Tr\cup X)$ and $p(\bm{f}|t,Tr\cup X)$. However, previous works have proved that if we do not balance the bias from the correlation between treatment and covariates, the model will tend to give biased prediction and thus can not give precise estimation of the treatment effect. A crucial bias in social media data is information cocoon: personalized recommendation systems will deliver user the contents that they are interested in. For example, it will deliver more anti-vaccine posts to anti-vaccine users because they are more likely to be interested in those contents. As a result, the anti-vaccine users will engage more with anti-vaccine posts. 

To address the above issue, following previous works in neural counterfactual prediction, we apply domain adversarial training to learn a representation $h$ that is invariant to such a bias\cite{bica2020estimating}. More specifically, we hope to learn a encoder $H$ such that for any two users with different history $X_1$ and $X_2$, $p(Tr|H(X_1))=p(Tr|H(X_2))$ for the same $Tr$. In other words, in the representation space, the probability that the two users interact with the same post at the same time should be same, which is the same as psychology experiments that divide the experimental and controlled groups randomly. To achieve this object, we apply adversarial training to remove the information about future treatment from the representation of covariates. More specifically, we additionally train a treatment predictor $\hat{p}(Tr|H(X))$ by modeling $\lambda_{tr}(t^{(e)}|h)$ and $p_{tr}(\bm{f}^{(e)}|t,h)$ with the encoding $h$ of the historical covariates $X$. However, between the treatment predictor and the encoder, we insert a \textbf{gradient reversal layer (GRL)}\cite{ganin2015unsupervised,raff2018gradient,bica2020estimating} to reverse the sign of the gradient. Thus, when we optimize the treatment predictor to maximize the likelihood of the observed treatment $Tr=(\bm{f}^{(e)},t^{(e)})$, the GRL will make the encoder to minimize the likelihood. This process leads to the following min-max game:
\begin{equation}
    \text{min}_{H}\text{max}_{\hat{p}}\mathbb{E}_{Tr,X\sim p(Tr,X)}\log\hat{p}(Tr|H(X))
\end{equation}
The following theorem (proof provided in Appendix \ref{theo:1}) provide theoretic guarantee that the above adversarial training reduce the bias introduced by treatment-covariate correlation (e.g. recommendation system and personal interest):
\begin{theorem}
Given the following min-max game: 
\begin{equation}
    \text{min}_{H}\text{max}_{\hat{p}}\mathbb{E}_{Tr,X\sim p(Tr,X)}\log\hat{p}(Tr|H(X)) 
\end{equation}
the min gamer's Nash balanced solution $H^*$, ensures for any $X_1,X_2$, the following equation holds:
\begin{equation}
    p(Tr|H^*(X_1)) = p(Tr|H^*(X_2))
\end{equation}
where $p$ denote the ground-truth conditional distribution of treatment given encoding.
\end{theorem}

\section{Experiments}

On real-world social media platforms, the ground truth causal effects of user engagement with posts, no matter misinformation post or information post, are unknown. To address the unknown ground truth causal effect, previous works of causality analysis evaluate their models on synthetic dataset. In this paper, following previous works, we evaluate the performance of our model and compare it with baselines on synthetic dataset. Then we apply our proposed method to evaluate the impact misinformation on a real-world social media data about COVID-19 vaccine collected from Twitter.

\subsection{Experiments on Synthetic Data}
\label{gen_inst}
\textbf{Synthetic Data Generation:} To simulate the real-world social media, we generate 15000 users and 120 post of news. Each user $i$ is represented with a hidden vector $u_i$, which correspond to the status of a social media user. Each piece news $n$ has two randomly generated feature vectors: a topic vector $v_{topic}(n)$ and an inherent influence vector $v_{in}(n)$. Each user has two kinds of activities: (1) engaging with one of the 120 news post and (2) posting a post with original contents. The chance that a user engage with a post is decided by $v_{in}(n)$ and $u_i$, simulating \textbf{information cocoons}. Engagement event with news post $n$ will change the hidden status $u$ of the user. The scale and direction of the change are decided by the current user status, the topic vector and the inherent influence vector jointly. 
For each user, the engagement events and the posting events are modeled through two Hawkes process respectively. Both Hawkes processes are influenced by user status $u$. Also, the feature of each posting event $\bm{f}$ is drawn from a distribution $P(\bm{f}|u,t)$ characterized by a random parameterized multi-layer perceptron (MLP) taking $(u,t)$ and random noises as input and output $\bm{f}$.  
Thus, the engagement with the news post will have causal effects on the two processes. Since we have all parameters of the model, we can calculate the ground-truth ITE defined in Eq. 7 for the synthetic dataset. The details of the data generation algorithm is included in the Appendix \ref{app:data_gen}. 

\textbf{Baselines:} To the best of our knowledge, the causality effect on temporal user behaviour from misinformation is not explored by previous works. Thus, we lack well-established baselines for this specific task. To address this issue, we select some baselines from previous works on temporal point process and temporal causal inference and extend them to adapt our setting. \textbf{FullyNN} \cite{omi2019fully}  is a non-causal neural temporal point process that predict the user future behaviours without considering causal effect. We select  it because has the same neural architecture as our model. It can also be regarded as our model's variant \textbf{w/o adversarial balancing}. \textbf{Neural-CIP}\footnote{Because the authors did not opensource the code of original CIP and one important precious work that is crucial for CIP, we implement a version of CIP that apply neural network rather than graphical causal model.} is an extension of CIP \cite{gao2021causal}, which aims at discovering causal effect of event pairs in temporal point process. We further compare our model with an ablation variant: \textbf{CNTPP-VAE}. It replace our GMM-based decoder with a Variational Auto-Encoder \cite{Kingma2014Auto}. Since VAE does not have a closed form solution of feature expect, we report the results applying sampling and approximation separately.

In this work, we will evaluate the proposed model in two aspects:

\begin{table}[!tb]
\caption{Estimation Error to the ground-truth ITE}
    \label{tab:my_label1}
    \vspace{0.3cm}
    \centering
    \begin{tabular}{|c|c|c|c|c|}
    \hline
         Method& Accuracy $\uparrow$ & RAE $\downarrow$ & RRSE $\downarrow$ & Decoder Inference Time\\
         \hline
         FullyNN & 73.0\% &0.865 & 0.901 &7.13ms \\
         CNTPP-VAE (Approximation)  & 85.9\% & 0.279& 0.503 &\textbf{4.05ms} \\
         CNTPP-VAE (Sampling) & 87.8\% &0.237& 0.454 & 29.34ms \\
         CNTPP(Ours)  & \textbf{88.0\%} & \textbf{0.234}& \textbf{0.448}  & 7.12ms \\
         \hline
    \end{tabular}
\end{table}

\begin{wraptable}{r}{7.8cm}
    \caption{Causal Effect Inference}
    \label{tab:my_label2}
    \centering
    \begin{tabular}{|c|c|c|}
    \hline
         Method& MatDis$\downarrow$ & LinCor$\uparrow$\\ 
         \hline
         Neural-CIP  & 0.90& 0.04\\
         FullyNN  & 0.93& 0.236\\
         CNTPP-VAE (Approximation)  & 0.84& 0.303\\
         CNTPP-VAE (Sampling) & \textbf{0.76}& 0.287 \\
         CNTPP (Ours)  & 0.77& \textbf{0.310}\\
         \hline
    \end{tabular}
    
\end{wraptable}

\textbf{ITE Estimation}: we will evaluate the model by comparing ITE estimated by the model and the ground-truth ITE. We report three metrics: \textbf{Accuracy} (the model need to correctly predict whether the engagement increase or decrease each dimension of the expected average feature), Relative Absolute Error (\textbf{RAE}) and Relative Root Square Error (\textbf{RRSE}). In addition, we also report the inference time of our model to reflect our model's efficiency.

As shown in Table~\ref{tab:my_label1}, our model establishes new state-of-the-art on all three quantitative metrics. This means that our model can fully utilize the causal information from the multivariates point process for unbiased treatment effect estimation. 
In particular, the model with causal analysis outperforms the model with direct neural network prediction (FullyNN), which leads to results that are not causally related. 
Simultaneously, our model outperforms all baselines in all three metrics of estimation precision. Although CRN-VAE incorporated with sampling method can achieve a performance very close to us, it spends substantially longer inference time.

\textbf{Causal Effect Inference}: CIP defines the treatment effect in a way different from our model. Thus ITE estimation experiment is not fair for it. For a fair comparison, and also to further prove that our model can achieve unbiased estimation, we use all the models to predict the ATE of each news post. Then we evaluate the correlation between the learnt ATEs and ground-truth average change of the news post to all users' hidden statuses. The more correlated the learnt ATE is, the better it reflect the inherent causal effect of the engagement on users. To evaluate the correlation, we apply the following two metrics: \textbf{MatDis} and \textbf{LinCor}. MatDis evaluates the similarity between the ATE-Distance matrix and Hidden-Status-Distance matrix. LinCor evaluates the linear correlation between the learnt ATE and the ground-truth average hidden status change. Details of the two metrics can be found in Appendix \ref{app:synthetic_metric}. From Table~\ref{tab:my_label2}, the ATE of our model best reflect the ground-truth average change of the news post on the simulated data for both two evaluation metrics. This suggests that our model has the potential to discover the influence of misinformation on social media users' hidden status, e.g. interest and idea.

\subsection{Experiments on Real World Data}
\begin{table}[!tb]
    \caption{Comparison of (normalized) Average Sum of Distances with different methods on the real-world dataset. This metric reflect how well the a group of data points is clustered.}
    \label{tab:my_label4}
    \vspace{0.5cm}
    \centering
    \begin{tabular}{|c|c|c|c|}
    \hline
         Methods& ASD$\downarrow$ &$\text{ASD}_{in}\downarrow$ &$\text{ASD}_{mis}\downarrow$ \\
         \hline
         Event Feature& 0.123&0.123&0.122 \\
         FullyNN &0.073 &0.069 &0.072 \\
         CNTPP (Ours) & \textbf{0.045} &\textbf{0.042} &\textbf{0.044}\\
         \hline
    \end{tabular}
    
\end{table}

In this section, we apply our proposed model on the Twitter dataset to estimate misinformation impact on social media scenario. We apply the data set collected in ~\cite{zhang2021vigdet,sharma2021covid}, including a total of 16,9008 tweets with labels from 24,192 users over a 5-month period from 2020/12/09 to 2021/04/24. Notably, we focus on understand how the tweets that users retweeted influence their behavior of posting original tweets. For each post, its feature $\bm{f}$ includes: text representation (extracted with a pre-trained BERT), sentiment score and subjectivity score. We discover the following two phenomenons with our model.

\begin{figure}[htbp]
\centering
\subfigure[Impact on sentiments and subjectivity.]{
\begin{minipage}[t]{0.48\linewidth}
\centering
\includegraphics[height=4.0cm]{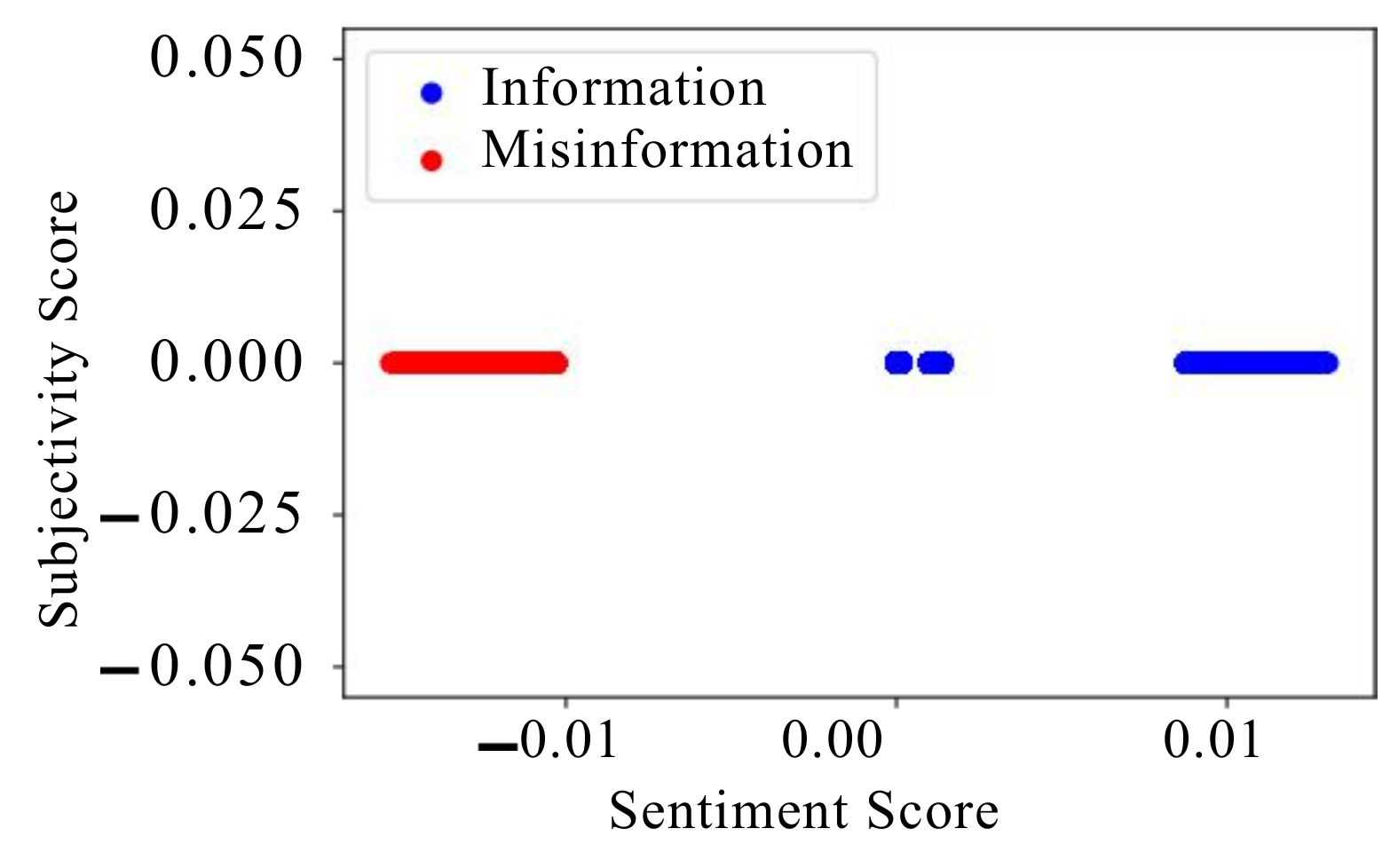}
\label{analysis_fin1}
\end{minipage}
}
\subfigure[ Impact on text representation.]{
\begin{minipage}[t]{0.48\linewidth}
\centering
\includegraphics[height=4.0cm]{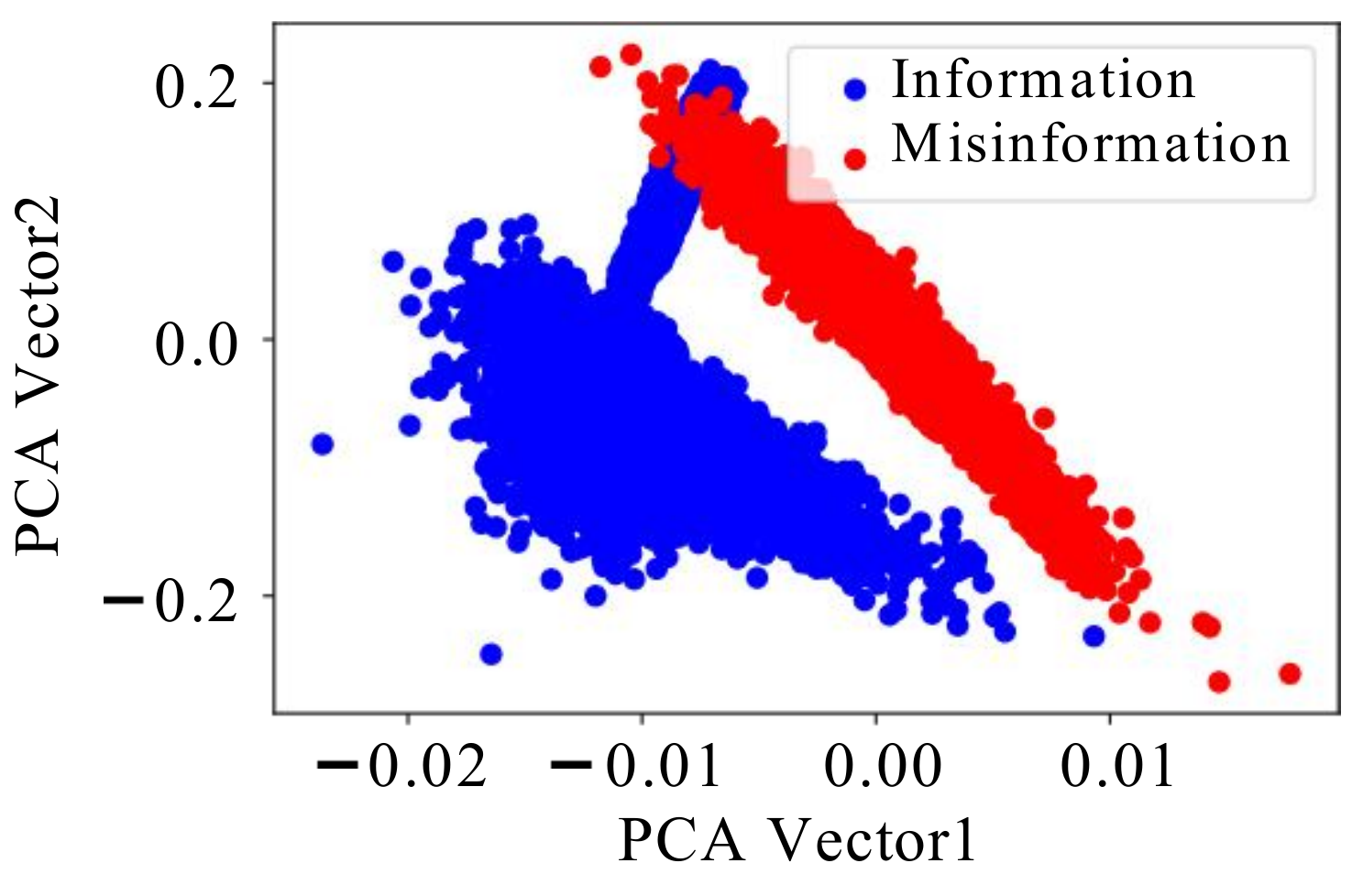}
\label{analysis_fin2}
\end{minipage}
}
\caption{Analysis on real world social media  data}
\label{analysis_fin3}
\end{figure}

\textbf{Identifiability between misinformation and information in influencing people's narratives}: 
For the desired outcome, we analyze the distinguishability between "retweeting fake news" and "retweeting true news" events. 
More specifically, for each retweeting event, we use its ITE estimated with our model as feature (dimension reduced via PCA \cite{martinez2001pca}) and whether the content is information or misinformation as label. As shown on Figure~\ref{analysis_fin2}, we can verify the identifiability of our proposed method as the treatment effect of two types of the news are substantially different. This discovery not only supports that misinformation and information influence people's behaviour in different ways, but also provides us with a new paradigm to detect fake news. We also calculate Normalized Averaged Sum of Distances (NASD, details in Appendix~\ref{app:real_metric}) for information cluster, misinformation cluster and their joint set. The lower these metrics are, the better that information and misinformation are distinguished. The comparison of our model against FullyNN and event features on this metric is shown in Table \ref{tab:my_label4}. As we can see, the ITE learnt by our model can identify information and misinformation better than the baselines.

\textbf{Misinformation is hurting people's subjective emotion related to COVID vaccine}: 
To understand the influence of misinformation in a more intuitive way, we analyze the impact of retweeting events on users' average sentiment score and subjectivity score in the future. The higher subjectivity the content gets, the more personal opinions rather than factual information text contains. Then, we use the proposed model to generate the estimated ITE for each retweeting event and plot the results of fake news and real news. As shown on Figure~\ref{analysis_fin1} (x-axis for sentiment score and y-axis for subjectivity score), we find that both information and misinformation do not substantially influence the users' subjectivity. However, information tends to make people optimistic about vaccines (true news increases the sentiment score), while fake news tends to make people feel negative about vaccines. This discovery strongly supports the hypothesis that misinformation is hurting people's subjective emotion toward COVID-19 vaccines, and suggests that misinformation could be causally responsible for vaccine hesitancy.

\section{Broader Impact and Limitations}
The predicted ITE scores of our model can bring impacts from two perspectives: \textbf{misinformation mitigation} and \textbf{misinformation research}. First, the predicted ITE scores help platforms allocate resources better for more efficient and effective misinformation mitigation. Here, resources include a wide range of specific concepts, including but limited to the efforts of human verifiers, users’ capacity to accept and spread the contents for clarification, and so on. Second, our proposed model provides researchers with a data-driven algorithmic tool to bridge the research in user behavior modeling and misinformation. This tool can help researchers in different ways, e.g. providing researchers a set of potential misinformation factors that could influence user behaviours, understanding misinformation campaigns, which spread misinformation with specific topics or narratives to influence public opinions, and designing better evaluation metrics for fake news detection\footnote{We are thankful to our anonymous reviewers for the discussion on this issue.}. 

The proposed model also has some limitations. First, it mainly focus on the causal effect of engagement on posting. However, in real-world social media, there could be other impacts of misinformation, such as changing the user's preference of engagement, topics of interest and community identity~\cite{pmlr-v162-zhang22ab}. Also, due to the limitation of synthetic algorithm and the meta data in the real-world data, we did not consider that different types of engagement may have different impact strength. In addition, the real-world dataset experiment only consider one dataset related to COVID-19, which is a single topic dataset. Although our model does not prohibit from being generalized onto multi-topic datasets, e.g., PolitiFact~\cite{vo2020facts} and GossipCop~\cite{shu2017fake, shu2018fakenewsnet, shu2017exploiting}, how to verify the model performance and reliability on a multi-topic dataset is still questionable. These limitations provides a strong motivation for further exploration on this paper's topic in future works. Potential directions may include how to verify the model's reliability on multi-topic datasets, how to generate synthetic data with more details and how to model more causal effect in real-world social media.

\section{Conclusion}
In this paper, we propose a framework to describe the causal structure model and causal effect about how misinformation influence online user behaviours. We further design a neural temporal point process model to conduct unbiased estimation on the causal effect in a data-driven approach. Experiments on synthetic dataset verify the effectiveness and efficiency of our model. We further apply our model on real-world dataset from Twitter and recognize identifiable causal effect of misinformation. The experiment results suggests that the misinformation about COVID-19 vaccine is hurting people's subjective attitudes toward vaccines.
However, it is also noticeable that our model is a statistical machine learning model. Consequently, all of its estimation can only be regarded as a reference rather than judgement. Also, misinformation campaigns could use the proposed approach to direct their editors to write more impactful fake news. A probable strategy to address this potential problem is to require social media platforms to raise necessary alerts to those suspicious articles that seems to be optimized. We discussed the strategy to detect such articles in the Checklist.

\section{Acknowledgement and Funding Disclosure}
This work is supported by NSF Research Grant IIS-2226087. Views and conclusions are of the authors and should not be interpreted as representing the social policies of the funding agency, or U.S. Government. Yizhou Zhang and Defu Cao are also partly supported by the Annenberg Fellowship of the University of Southern California. We sincerely appreciate the feedback, comments and suggestions from our anonymous reviewers.

\bibliography{ref}

\begin{thebibliography}{10}

\bibitem{aalen2020time}
Odd~O Aalen, Mats~J Stensrud, Vanessa Didelez, Rhian Daniel, Kjetil
  R{\o}ysland, and Susanne Strohmaier.
\newblock Time-dependent mediators in survival analysis: Modeling direct and
  indirect effects with the additive hazards model.
\newblock {\em Biometrical Journal}, 62(3):532--549, 2020.

\bibitem{bahadori2021debiasing}
Mohammad~Taha Bahadori and David Heckerman.
\newblock Debiasing concept-based explanations with causal analysis.
\newblock In {\em International Conference on Learning Representations}, 2021.

\bibitem{Bellot2021PolicyAU}
Alexis Bellot and Mihaela van~der Schaar.
\newblock Policy analysis using synthetic controls in continuous-time.
\newblock In {\em ICML}, 2021.

\bibitem{bica2020estimating}
Ioana Bica, Ahmed~M Alaa, James Jordon, and Mihaela van~der Schaar.
\newblock Estimating counterfactual treatment outcomes over time through
  adversarially balanced representations.
\newblock {\em International Conference on Learning Representations}, 2020.

\bibitem{campbell:01}
N.~Campbell.
\newblock The study of discontinuous phenomena.
\newblock {\em Proc Cambr. Phil. Soc}, vol. 15:pp. 117--136, 1909.

\bibitem{cao2020spectral}
Defu Cao, Yujing Wang, Juanyong Duan, Ce~Zhang, Xia Zhu, Congrui Huang, Yunhai
  Tong, Bixiong Xu, Jing Bai, Jie Tong, et~al.
\newblock Spectral temporal graph neural network for multivariate time-series
  forecasting.
\newblock {\em Advances in neural information processing systems},
  33:17766--17778, 2020.

\bibitem{chen2020neural}
Ricky~TQ Chen, Brandon Amos, and Maximilian Nickel.
\newblock Neural spatio-temporal point processes.
\newblock {\em arXiv preprint arXiv:2011.04583}, 2020.

\bibitem{cheng2021causal}
Lu~Cheng, Ruocheng Guo, Kai Shu, and Huan Liu.
\newblock Causal understanding of fake news dissemination on social media.
\newblock In {\em Proceedings of the 27th ACM SIGKDD Conference on Knowledge
  Discovery \& Data Mining}, pages 148--157, 2021.

\bibitem{cheng2021causala}
Lu~Cheng, Ahmadreza Mosallanezhad, Paras Sheth, and Huan Liu.
\newblock Causal learning for socially responsible ai.
\newblock In {\em 30th International Joint Conference on Artificial
  Intelligence, IJCAI 2021}, pages 4374--4381. International Joint Conferences
  on Artificial Intelligence, 2021.

\bibitem{de2019temporal}
Abir De, Utkarsh Upadhyay, and Manuel Gomez-Rodriguez.
\newblock Temporal point processes.
\newblock {\em Technical report, Technical report, Saarland University}, 2019.

\bibitem{devlin2018bert}
Jacob Devlin, Ming-Wei Chang, Kenton Lee, and Kristina Toutanova.
\newblock {BERT}: Pre-training of deep bidirectional transformers for language
  understanding.
\newblock In {\em Proceedings of the 2019 Conference of the North {A}merican
  Chapter of the Association for Computational Linguistics: Human Language
  Technologies, Volume 1 (Long and Short Papers)}, pages 4171--4186,
  Minneapolis, Minnesota, June 2019. Association for Computational Linguistics.

\bibitem{du2016recurrent}
Nan Du, Hanjun Dai, Rakshit Trivedi, Utkarsh Upadhyay, Manuel Gomez-Rodriguez,
  and Le~Song.
\newblock Recurrent marked temporal point processes: Embedding event history to
  vector.
\newblock In {\em Proceedings of the 22nd ACM SIGKDD International Conference
  on Knowledge Discovery and Data Mining}, pages 1555--1564, 2016.

\bibitem{ganin2015unsupervised}
Yaroslav Ganin and Victor Lempitsky.
\newblock Unsupervised domain adaptation by backpropagation.
\newblock In {\em International conference on machine learning}, pages
  1180--1189. PMLR, 2015.

\bibitem{gao2021causal}
Tian Gao, Dharmashankar Subramanian, Debarun Bhattacharjya, Xiao Shou, Nicholas
  Mattei, and Kristin Bennett.
\newblock Causal inference for event pairs in multivariate point processes.
\newblock In A.~Beygelzimer, Y.~Dauphin, P.~Liang, and J.~Wortman Vaughan,
  editors, {\em Advances in Neural Information Processing Systems}, 2021.

\bibitem{jolley2014effects}
Daniel Jolley and Karen~M Douglas.
\newblock The effects of anti-vaccine conspiracy theories on vaccination
  intentions.
\newblock {\em PloS one}, 9(2):e89177, 2014.

\bibitem{kamra2021polsird}
Nitin Kamra, Yizhou Zhang, Sirisha Rambhatla, Chuizheng Meng, and Yan Liu.
\newblock Polsird: Modeling epidemic spread under intervention policies.
\newblock {\em Journal of Healthcare Informatics Research}, 5(3):231--248,
  2021.

\bibitem{Kingma2014Auto}
D.~P. Kingma and M.~Welling.
\newblock Auto-encoding variational bayes.
\newblock In {\em International Conference on Learning Representations (ICLR)},
  2014.

\bibitem{kricorian2022covid}
Katherine Kricorian, Rachel Civen, and Ozlem Equils.
\newblock Covid-19 vaccine hesitancy: Misinformation and perceptions of vaccine
  safety.
\newblock {\em Human Vaccines \& Immunotherapeutics}, 18(1):1950504, 2022.

\bibitem{loria2018textblob}
Steven Loria et~al.
\newblock textblob documentation.
\newblock {\em Release 0.15}, 2:269, 2018.

\bibitem{luceri2020detecting}
Luca Luceri, Silvia Giordano, and Emilio Ferrara.
\newblock Detecting troll behavior via inverse reinforcement learning: A case
  study of russian trolls in the 2016 us election.
\newblock In {\em Proceedings of the International AAAI Conference on Web and
  Social Media}, volume~14, pages 417--427, 2020.

\bibitem{martinez2001pca}
Aleix~M Martinez and Avinash~C Kak.
\newblock Pca versus lda.
\newblock {\em IEEE transactions on pattern analysis and machine intelligence},
  23(2):228--233, 2001.

\bibitem{mei2017neural}
Hongyuan Mei and Jason~M Eisner.
\newblock The neural hawkes process: A neurally self-modulating multivariate
  point process.
\newblock In {\em Advances in Neural Information Processing Systems}, pages
  6754--6764, 2017.

\bibitem{meng2022physics}
Chuizheng Meng, Sungyong Seo, Defu Cao, Sam Griesemer, and Yan Liu.
\newblock When physics meets machine learning: A survey of physics-informed
  machine learning.
\newblock {\em arXiv preprint arXiv:2203.16797}, 2022.

\bibitem{noorbakhsh2021counterfactual}
Kimia Noorbakhsh and Manuel~Gomez Rodriguez.
\newblock Counterfactual temporal point processes.
\newblock {\em arXiv preprint arXiv:2111.07603}, 2021.

\bibitem{omi2019fully}
Takahiro Omi, Kazuyuki Aihara, et~al.
\newblock Fully neural network based model for general temporal point
  processes.
\newblock {\em Advances in neural information processing systems}, 32, 2019.

\bibitem{pearl2009causality}
Judea Pearl.
\newblock {\em Causality}.
\newblock Cambridge university press, 2009.

\bibitem{perez2017automatic}
Ver{\'o}nica P{\'e}rez-Rosas, Bennett Kleinberg, Alexandra Lefevre, and Rada
  Mihalcea.
\newblock Automatic detection of fake news.
\newblock In {\em Proceedings of the 27th International Conference on
  Computational Linguistics}, pages 3391--3401, Santa Fe, New Mexico, USA,
  August 2018. Association for Computational Linguistics.

\bibitem{raff2018gradient}
Edward Raff and Jared Sylvester.
\newblock Gradient reversal against discrimination: A fair neural network
  learning approach.
\newblock In {\em 2018 IEEE 5th International Conference on Data Science and
  Advanced Analytics (DSAA)}, pages 189--198, 2018.

\bibitem{reshi2022covid}
Aijaz~Ahmad Reshi, Furqan Rustam, Wajdi Aljedaani, Shabana Shafi, Abdulaziz
  Alhossan, Ziyad Alrabiah, Ajaz Ahmad, Hessa Alsuwailem, Thamer~A Almangour,
  Musaad~A Alshammari, et~al.
\newblock Covid-19 vaccination-related sentiments analysis: A case study using
  worldwide twitter dataset.
\newblock In {\em Healthcare}, volume~10, page 411. MDPI, 2022.

\bibitem{roysland2012counterfactual}
Kjetil R{\o}ysland.
\newblock Counterfactual analyses with graphical models based on local
  independence.
\newblock {\em The Annals of Statistics}, 40(4):2162--2194, 2012.

\bibitem{ryalen2020causal}
P{\aa}l~Christie Ryalen, Mats~Julius Stensrud, Sophie Foss{\aa}, and Kjetil
  R{\o}ysland.
\newblock Causal inference in continuous time: an example on prostate cancer
  therapy.
\newblock {\em Biostatistics}, 21(1):172--185, 2020.

\bibitem{salem2019fa}
Fatima K~Abu Salem, Roaa Al~Feel, Shady Elbassuoni, Mohamad Jaber, and May
  Farah.
\newblock Fa-kes: a fake news dataset around the syrian war.
\newblock In {\em Proceedings of the International AAAI Conference on Web and
  Social Media}, volume~13, pages 573--582, 2019.

\bibitem{samel2022learning}
Karan Samel, Zelin Zhao, Binghong Chen, Shuang Li, Dharmashankar Subramanian,
  Irfan Essa, and Le~Song.
\newblock Learning temporal rules from noisy timeseries data.
\newblock {\em arXiv preprint arXiv:2202.05403}, 2022.

\bibitem{sanaullah2022applications}
AR~Sanaullah, Anupam Das, Anik Das, Muhammad~Ashad Kabir, and Kai Shu.
\newblock Applications of machine learning for covid-19 misinformation: a
  systematic review.
\newblock {\em Social Network Analysis and Mining}, 12(1):1--34, 2022.

\bibitem{sharma2019combating}
Karishma Sharma, Feng Qian, He~Jiang, Natali Ruchansky, Ming Zhang, and Yan
  Liu.
\newblock Combating fake news: A survey on identification and mitigation
  techniques.
\newblock {\em ACM Transcations on Intelligent Systems and TEchnology}, 2019.

\bibitem{sharma2020coronavirus}
Karishma Sharma, Sungyong Seo, Chuizheng Meng, Sirisha Rambhatla, Aastha Dua,
  and Yan Liu.
\newblock Coronavirus on social media: Analyzing misinformation in twitter
  conversations.
\newblock {\em arXiv preprint arXiv:2003.12309}, 2020.

\bibitem{sharma2020identifying}
Karishma Sharma, Yizhou Zhang, Emilio Ferrara, and Yan Liu.
\newblock Identifying coordinated accounts on social media through hidden
  influence and group behaviours.
\newblock In {\em Proceedings of the 27th ACM SIGKDD Conference on Knowledge
  Discovery \& Data Mining}, pages 1441--1451, 2021.

\bibitem{sharma2021covid}
Karishma Sharma, Yizhou Zhang, and Yan Liu.
\newblock Covid-19 vaccine misinformation campaigns and social media
  narratives.
\newblock In {\em Proceedings of the International AAAI Conference on Web and
  Social Media}, volume~16, pages 920--931, 2022.

\bibitem{shu2018fakenewsnet}
Kai Shu, Deepak Mahudeswaran, Suhang Wang, Dongwon Lee, and Huan Liu.
\newblock Fakenewsnet: A data repository with news content, social context, and
  spatiotemporal information for studying fake news on social media.
\newblock {\em Big data}, 8 3:171--188, 2020.

\bibitem{shu2017fake}
Kai Shu, Amy Sliva, Suhang Wang, Jiliang Tang, and Huan Liu.
\newblock Fake news detection on social media: A data mining perspective.
\newblock {\em ACM SIGKDD Explorations Newsletter}, 19(1):22--36, 2017.

\bibitem{shu2017exploiting}
Kai Shu, Suhang Wang, and Huan Liu.
\newblock Exploiting tri-relationship for fake news detection.
\newblock {\em arXiv preprint arXiv:1712.07709}, 2017.

\bibitem{van2015conspiracy}
Sander Van~der Linden.
\newblock The conspiracy-effect: Exposure to conspiracy theories (about global
  warming) decreases pro-social behavior and science acceptance.
\newblock {\em Personality and Individual Differences}, 87:171--173, 2015.

\bibitem{vo2020facts}
Nguyen Vo and Kyumin Lee.
\newblock Where are the facts? searching for fact-checked information to
  alleviate the spread of fake news.
\newblock In {\em Proceedings of the 2020 Conference on Empirical Methods in
  Natural Language Processing (EMNLP 2020)}, 2020.

\bibitem{wang2020csan}
Qi~Wang, Guangyin Jin, Xia Zhao, Yanghe Feng, and Jincai Huang.
\newblock Csan: A neural network benchmark model for crime forecasting in
  spatio-temporal scale.
\newblock {\em Knowledge-Based Systems}, 189:105120, 2020.

\bibitem{wilson2020social}
Steven~Lloyd Wilson and Charles Wiysonge.
\newblock Social media and vaccine hesitancy.
\newblock {\em BMJ Global Health}, 5(10):e004206, 2020.

\bibitem{ZEEVI199799}
Assaf~J. Zeevi and Ronny Meir.
\newblock Density estimation through convex combinations of densities:
  Approximation and estimation bounds.
\newblock {\em Neural Networks}, 10(1):99--109, 1997.

\bibitem{pmlr-v162-zhang22ab}
Junzhe Zhang, Jin Tian, and Elias Bareinboim.
\newblock Partial counterfactual identification from observational and
  experimental data.
\newblock In Kamalika Chaudhuri, Stefanie Jegelka, Le~Song, Csaba Szepesvari,
  Gang Niu, and Sivan Sabato, editors, {\em Proceedings of the 39th
  International Conference on Machine Learning}, volume 162 of {\em Proceedings
  of Machine Learning Research}, pages 26548--26558. PMLR, 17--23 Jul 2022.

\bibitem{zhangself}
Qiang Zhang, Aldo Lipani, Omer Kirnap, and Emine Yilmaz.
\newblock Self-attentive hawkes process.
\newblock {\em ICML}, 2020.

\bibitem{zhang2021vigdet}
Yizhou Zhang, Karishma Sharma, and Yan Liu.
\newblock Vigdet: Knowledge informed neural temporal point process for
  coordination detection on social media.
\newblock {\em Advances in Neural Information Processing Systems}, 34, 2021.

\bibitem{zuiderveen2016should}
Frederik Zuiderveen~Borgesius, Damian Trilling, Judith M{\"o}ller, Bal{\'a}zs
  Bod{\'o}, Claes~H De~Vreese, and Natali Helberger.
\newblock Should we worry about filter bubbles?
\newblock {\em Internet Policy Review. Journal on Internet Regulation}, 5(1),
  2016.

\bibitem{zuo2020transformer}
Simiao Zuo, Haoming Jiang, Zichong Li, Tuo Zhao, and Hongyuan Zha.
\newblock Transformer hawkes process.
\newblock {\em NeurIPS}, 2020.

\bibitem{zuo2022differentially}
Simiao Zuo, Tianyi Liu, Tuo Zhao, and Hongyuan Zha.
\newblock Differentially private estimation of hawkes process.
\newblock {\em arXiv preprint arXiv:2209.07303}, 2022.

\end{thebibliography}
\bibliographystyle{plain}
\clearpage

\section*{Checklist}

\begin{enumerate}

\item For all authors...
\begin{enumerate}
  \item Do the main claims made in the abstract and introduction accurately reflect the paper's contributions and scope?
    \answerYes{}
  \item Did you describe the limitations of your work?
    \answerYes{\textcolor{black}{In the Table 1 and 2, we honestly present the metrics under which the baselines outperform us. Also, in the conclusion section, we emphasize that this work is a statistical machine learning model. Thus, it can only work as a reference rather than judgement.}}
  \item Did you discuss any potential negative societal impacts of your work?
    \answerYes{See the conclusion. The detection of the fake news from misinformation campaigns mentioned in the conclusion section has multiple potential solutions. One solution is to use anomaly detection models to detect articles with extremely high ITE or ATE scores. Another probable solution is based on gradient and optimization. The process of guiding editors with the proposed model is equivalent to generating texts under a regularization to maximize the ITE or ATE scores estimated by the proposed model. Thus, the "optimized" articles will be closed to an optimal points, where gradients of the ITE or ATE scores with respect to the article feature is statistically lower than other points. Thus, the platforms can detect such articles by computing the article embedding's distance to the closest local maximum or gradient scale.}
  \item Have you read the ethics review guidelines and ensured that your paper conforms to them?
    \answerYes{}
\end{enumerate}

\item If you are including theoretical results...
\begin{enumerate}
  \item Did you state the full set of assumptions of all theoretical results?
    \answerYes{}
        \item Did you include complete proofs of all theoretical results?
    \answerYes{See the appendix in the supplementary materials}
\end{enumerate}

\item If you ran experiments...
\begin{enumerate}
  \item Did you include the code, data, and instructions needed to reproduce the main experimental results (either in the supplemental material or as a URL)?
    \answerYes{See the zip file in the supplementary materials.}
  \item Did you specify all the training details (e.g., data splits, hyperparameters, how they were chosen)?
    \answerYes{Due to the page limits, we put some them in the appendix in the supplementary materials.}
        \item Did you report error bars (e.g., with respect to the random seed after running experiments multiple times)?
    \answerYes{In the appendix.}
        \item Did you include the total amount of compute and the type of resources used (e.g., type of GPUs, internal cluster, or cloud provider)?
    \answerYes{Due to the page limits, we put some them in the appendix in the supplementary materials.}
\end{enumerate}

\item If you are using existing assets (e.g., code, data, models) or curating/releasing new assets...
\begin{enumerate}
  \item If your work uses existing assets, did you cite the creators?
    \answerYes{}
  \item Did you mention the license of the assets?
    \answerNo{The licenses are included in those assets on Github.}
  \item Did you include any new assets either in the supplemental material or as a URL?
    \answerYes{}
  \item Did you discuss whether and how consent was obtained from people whose data you're using/curating?
    \answerNA{}
  \item Did you discuss whether the data you are using/curating contains personally identifiable information or offensive content?
    \answerYes{In the dataset, we only preserve the extracted features and remove those personally identifiable information or specific text contents.}
\end{enumerate}

\item If you used crowdsourcing or conducted research with human subjects...
\begin{enumerate}
  \item Did you include the full text of instructions given to participants and screenshots, if applicable?
    \answerNA{}
  \item Did you describe any potential participant risks, with links to Institutional Review Board (IRB) approvals, if applicable?
    \answerNA{}
  \item Did you include the estimated hourly wage paid to participants and the total amount spent on participant compensation?
    \answerNA{}
\end{enumerate}

\end{enumerate}
\clearpage
\appendix
\newtheorem{lemma}{Lemma}
\section{Theoretic Results}
\setcounter{theorem}{0}
\subsection{Proof of the theorems}
\label{theo:1}
\begin{theorem}
For a user $u$, if the intensity function $\lambda(\bm{f},t|Tr\cup X)$ is known, then we have:
\begin{equation}
    \mu(t,t+T,\lambda_1,Tr\cup X) = \int_{\text{sup}(\bm{f})}d\bm{f}\int_t^{t+T}\lambda(\bm{f},t|Tr\cup X)dt
\end{equation}
\begin{equation}
    \phi(t,t+T,\lambda_1,Tr\cup X) = \int_{\text{sup}(\bm{f})}\bm{f}d\bm{f}\int_t^{t+T}\lambda(\bm{f},t|Tr\cup X)dt
\end{equation}
\end{theorem}
\begin{proof}
The first equation can be trivially proved by replacing the $\bm{F}$ in the definition of $\lambda$ with the support set. The second one can be proved with the Campbell's Theorem \cite{campbell:01}: 

\begin{lemma}
Campbell's Theorem (Campbell, 1909): For a point process $S$, and a measurable function $f:\mathbb{R}^d\rightarrow\mathbb{R}^{d'}$:
\begin{equation}
    \mathbb{E}[\sum_{x\in S}f(x)]=\int_{\mathbb{R}^d}f(x)\mathbb{E}(N(x,x+dx))
\end{equation}
\end{lemma}
Note that the above theorem is for all point process. For the specific scenario of temporal point process, we also need to include the time range into consideration:
\begin{equation}
    \mathbb{E}[\sum_{x\in S[t_1:t_2]}f(x)]=\int_{\mathbb{R}^d}f(x)\mathbb{E}(N(x,x+dx,t_1,t_2))
\end{equation}
From the definition of $\phi$, we know that the $f(x)=x$. Recall the definition of intensity function (Equation. 1 in the main content):
\begin{equation}
    \mathbb{E}(N(\bm{F},T_1,T_2)|S_h)=\int_{\bm{F}}d\bm{f}\int_{T_1}^{T_2}\lambda(\bm{f},t|S_h)dt
    \label{eq:diff}
\end{equation}
We can convert this integral to a derivative formula:
\begin{equation}
    \mathbb{E}(N(\bm{f},\bm{f}+d\bm{f},T_1,T_2)|S_h)=d\bm{f}\int_{T_1}^{T_2}\lambda(\bm{f},t|S_h)dt
\end{equation}
With the above formula and Campbell's theorem, we have:
\begin{equation}
    \phi(t,t+T,\lambda_1,Tr\cup X) = \int_{\text{sup}(\bm{f})}\bm{f}d\bm{f}\int_t^{t+T}\lambda(\bm{f},t|Tr\cup X)dt
\end{equation}
\end{proof}

\begin{theorem}
Given the following min-max game: 
\begin{equation}
    \text{min}_{H}\text{max}_{\hat{p}}\mathbb{E}_{Tr,X\sim p(Tr,X)}\log\hat{p}(Tr|H(X)) 
\end{equation}
the min gamer's Nash balanced solution $H^*$, ensures for any $X_1,X_2$, the following equation holds:
\begin{equation}
    p(Tr|H^*(X_1)) = p(Tr|H^*(X_2))
\end{equation}
where $p$ denote the ground-truth conditional distribution of treatment given encoding.
\end{theorem}
\begin{proof}
For the objective of the max gamer, we have:
\begin{equation}
\begin{aligned}
    \mathbb{E}_{Tr,X\sim p(Tr,X)}\log\hat{p}(Tr|H(X))=&\mathbb{E}_{Tr,H(X)\sim p(Tr,H(X))}\log\hat{p}(Tr|H(X))\\
    =&\mathbb{E}_{H(X)\sim p(H(X))}\mathbb{E}_{Tr\sim p(Tr|H(X))}\log\hat{p}(Tr|H(X))\\
    =&\mathbb{E}_{H(X)\sim p(H(X))}-KL(p(Tr|H(X)),\hat{p}(Tr|H(X))\\
    &-Q(p(Tr|H(X)))\\
\end{aligned}
\end{equation}
where $Q(p(Tr|H(X)))$ is the entropy of $p(Tr|H(X))$.
Note that for a fixed $H$, $Q(p(Tr|H(X)))$ is a constant. Thus, the max objective is the same as minimizing the KL-divergence between $p(Tr|H(X))$ and $\hat{p}(Tr|H(X)$ for every $X$:
\begin{equation}
    \text{max}_{\hat{p}}\mathbb{E}_{Tr,X\sim p(Tr,X)}\log\hat{p}(Tr|H(X))\iff \text{min}_{\hat{p}}\mathbb{E}_{H(X)\sim p(H(X))}KL(p(Tr|H(X)),\hat{p}(Tr|H(X))
\end{equation}
It is known that the KL-divergence is minimized if and only if $\hat{p}(Tr|H(X)$ perfectly fit $p(Tr|H(X))$ at each point. Thus, the balance solution of the max gamer is to learn $\hat{p}(Tr|H(X))=p(Tr|H(X))$. Meanwhile, if there are two $X_1$ and $X_2$ such that $p(Tr|H(X_1))\not=p(Tr|H(X_2))$, then the min gamer can always update the $H$ by swapping $H(X_1)$ and $H(X_2)$ while maintaining other points the same. This is doable when $H$ is a universal function approximator. As a result, the KL-divergence will increase, which violates the definition of Nash balance. Therefore, at the balance point for any $X_1$ and $X_2$, $p(Tr|H(X_1))=p(Tr|H(X_2))$
\end{proof}
\section{Details of the Experiments}
\subsection{Data and Code}
Together with this appendix, we also submit the synthetic data and necessary code. However, the real-world data (from Twitter) is too huge for OpenReview platform. Thus, we can not submit it right now. We will opensource a version of the data with anonymous processing to protect privacies after this paper is accepted.
\subsection{Synthetic Dataset}
\label{app:data_gen}
\begin{figure}
    \centering
    \includegraphics[width=\textwidth]{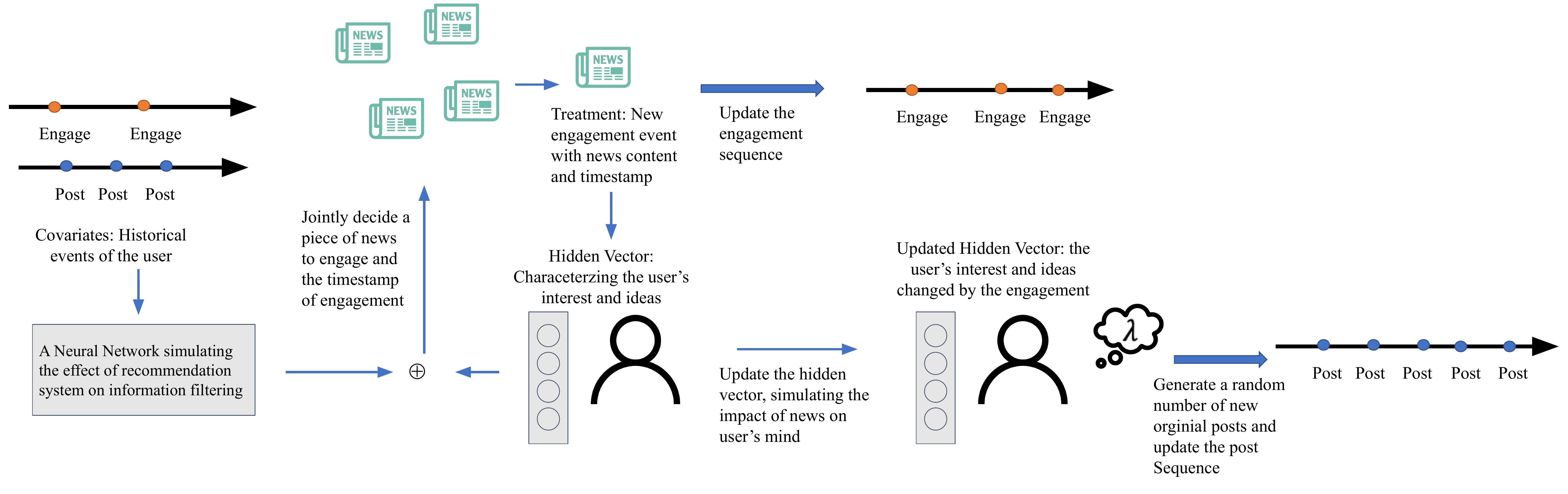}
    \caption{The general idea of generating the synthetic dataset.}
    \label{fig:syn}
\end{figure}

In this section, we introduce the how to generate the synthetic data in details. We present the basic ideas in Figure \ref{fig:syn}. To generate the data, we design a system to simulate the social media activities and events in real world. The basic elements in this system are \textbf{users} and \textbf{news}.
\begin{itemize}
    \item \textbf{User}: Each user $i$ is represented with a \textbf{hidden} vector $u_i$, which represent the hidden status of a social media user, such as interests, ideas and political trends in reality. If a user engage with a piece of news, then this user's hidden vector will change accordingly. We randomly assign each user with an randomly initialized vector to simulate the fact that users may hold different ideas and interests before starting to use social media.
    \item \textbf{News}: Each piece of news $n$ is characterized by two randomly generated \textbf{hidden} feature vectors: a topic vector $v_{topic}(n)$ and an inherent influence vector $v_{in}(n)$. These two vectors decide which users will be attracted by the news and how they will be influenced by the news.
\end{itemize} 

In this system, a user has two kinds of activities: (1) engaging with a piece of news and (2) posting a post with original contents. These activities are characterized by two temporal point process respectively.
For the process of engaging with a piece of news, the intensity function of a user's ($i$) engagement with a specific piece of news $n$ is defined as:
\begin{equation}
    \lambda_{e}(n,t|u_i,S_h^{(e)},S_h^{(g)}) = v_{topic}(n)\cdot u_i + \sum_{(\bm{f}_j,t_j)\in S_h^{(e)}\cup S_h^{(g)}}\exp(t_j-t)NN_{rec}(\bm{f}_j, v_{topic}(n))
\end{equation}
where $S_h^{(e)}$ and $S_h^{(g)}$ denotes the historical sequences of a user's engagement and posting activities respectively. In this equation, the first term models the consistency of news topic and user interest and the $NN_{rec}(\cdot,\cdot)$ in the second term is a random parameterized neural network simulating the recommendation system that recommend contents based on past history. $\exp(t_j-t)$ make sure that the influence of a history event decrease as time goes by, simulating the property that recommendation system usually give higher attention to recent events. 

Similarly, the intensity function of a user's ($i$) process generating original posts is defined as:
\begin{equation}
    \lambda_{g}(t|u_i,S_h^{(e)},S_h^{(g)}) = u_i\cdot u_i + \sum_{(\bm{f}_j,t_j)\in S_h^{(e)}\cup S_h^{(g)}}\exp(t_j-t)NN_{post}(\bm{f}_j, u_i)
\end{equation}
where $u_i\cdot u_i$ models the user's reliance to social media and the second term reflect the influence of history events to future event. 

In real world, the news topic, influence on user mind and user's mind status are not observable. Instead we can only observe the tweet content text features, such as the representation extracted with a pre-trained language model. To simulate this fact, we use two neural networks $NN_{news}$ and $NN_{user}$ with random parameters to project the vectors to observable features:
\begin{equation}
    \bm{f}_{news}(n) = NN_{news}(v_{topic}(n),v_{in}(n)),\quad\bm{f}_{user}(i,t) = NN_{user}(u_i,t)
\end{equation}
where $\bm{f}_{news}(n)$ is the observable content feature of news $n$ and $\bm{f}_{user}(i,t)$ is the observable content feature of the post generated by user $i$ at time $t$. $\bm{f}_{news}(n)$ does not change over time because once a piece of news is reported, its content does not change significantly. Note that a normal user's activity is usually uncertain. To simulate such uncertainty, in each hidden layer of $NN_{user}$, we add a random noise. Since the two networks are with random parameters, the projection will be highly non-linear and thus brings enough complexity on the learning of causal impact.

The overall algorithm is shown in Algorithm \ref{alg}.

\begin{algorithm}

\caption{Synthetic Data Generation}
\begin{algorithmic}[1]
\STATE Set the ending time $T$. \COMMENT{The simulation will end at time $T$}
\STATE Randomly generated user hidden representations and news features and a posting time of each news.\COMMENT{Each piece of news will only be engaged with after it is posted.}
\FOR {news $n$}
\STATE $\bm{f}_{news}(n) = NN_{news}(v_{topic}(n),v_{in}(n))$

\ENDFOR
\FOR {user $i$}
\STATE $t_{end}=0$
\STATE $S_h^{(e)},S_h^{(g)}=\emptyset,\emptyset$
\REPEAT
\STATE Draw a posting event timestamp $t_{g}$ based on intensity function $\lambda_{g}(t|u_i,S_h^{(e)},S_h^{(g)})$
\STATE $t_{cur} = t_{g}$
\STATE $\bm{f}_{cur} = \bm{f}_{user}(i,t_{g})$
\STATE $flag='generating'$
\FOR {news $n$}
\IF{$t_p(n) \geq t_{cur}$}
\STATE Continue.
\ENDIF
\STATE Draw an event timestamp $t_e$ based on intensity function $\lambda_{e}(n,t|u_i,S_h^{(e)},S_h^{(g)})$
\IF{$t_e < t_{cur}$}
\STATE $t_{cur} = t_e$
\STATE $\bm{f}_{cur} = \bm{f}_{news}(n)$
\STATE $flag='engagement'$
\STATE $n_{cur}=n$
\ENDIF
\ENDFOR
\STATE $t_{end}=t_{cur}$
\IF{$flag$ is $'engagement'$}
\STATE $S_h^{(e)}=S_h^{(e)}\cup (\bm{f}_{cur},t_{cur})$
\STATE $u_i=u_i+NN_{scale}(v_{topic}(n_{cur}),u_i)v_{in}(n_{cur})$ \COMMENT{Update the user status. $NN_{scale}$ is a neural network with random parameters to adjust news impact scale.}
\ELSE
\STATE $S_h^{(g)}=S_h^{(g)}\cup (\bm{f}_{cur},t_{cur})$
\ENDIF
\UNTIL{$t_{end}\geq T$}
\ENDFOR
\end{algorithmic}
\label{alg}
\end{algorithm}

\subsection{Synthetic Data Experiments}

We remove the users with only one event and then acquire around 13000 accounts. Then we split the training set, validation set and test set according to 8000/2000/3000 to train the proposed model. Specifically, we adopt the Adam optimizer with learning rate $1e-3$ and the max epoch is set to be 500 with batch size 128. In addition, we add the dropout with 0.1 rate. To evaluate their decoder inference time, we first forward the data into the encoder to acquire the sequence embeddings and then compute the time that the decoder requires to predict the ITE. The decoder inference time with an error bar is presented in Table \ref{tab:my_label3}  
\begin{table}[htb]
\caption{Inference Time with Error Bar}
    \label{tab:my_label3}
    \vspace{0.2cm}
    \centering
    \begin{tabular}{|c|c|}
    
    \hline
         Method& Inference Time\\
         \hline
         FullyNN  &7.13$\pm$0.2 ms \\
         CNTPP-VAE (Approximation) &\textbf{4.05$\pm$0.1ms} \\
         CNTPP-VAE (Sampling) & 29.34$\pm$ 2.3ms \\
         CNTPP(Ours)   & 7.12$\pm$ 0.2ms \\
         \hline
    \end{tabular}
\end{table}

\subsection{Evaluation Metrics of Synthetic Data}
\label{app:synthetic_metric}

In synthetic data experiment, we applied MatDis and LinCor to evaluate the correlation between the learnt ATE and the ground-truth user hidden statuses change. MatDis evaluate the similarity between the ATE-Distance matrix and Hidden-Status-Distance matrix. More specifically, we first compute the post pairwise distances based on the learnt ATE and ground-truth hidden status respectively to acquire the ATE-Distance matrix and Hidden-Status-Distance matrix. Then we sort the posts to the same order in both matrices. After that we normalize the two matrices and compute their L2 distance. As for LinCor, we train a linear regression model to predict the ground-truth average hidden status change of a post given its learnt ATE. Then we evaluate the score of the linear model.

\subsection{Real World Dataset}
To collect the Twitter datset related to COVID-19 vaccines, we use the tracked keywords including vaccine, Pfizer, BioNTech, Moderna, Janssen, AstraZeneca, Sinopharm to filtering all tweets via the streaming Twitter API which returns 1\% sample of all tweets. We collect the tweets just before Pfizer-BioNTech and Moderna were approved by the FDA for Emergency Use Authorization (EUA) from Dec 9, 2020 - April 24, 2021. After that, we only use the tweets with labels of 'MisInformation' and 'Information', including a total of 169,008 tweets from 24,192 users.

\subsection{Real World Data Experiments}

For the impact on sentiments and subjectivity experiments, we use first use TextBlob sentiment analyse tool~\cite{loria2018textblob, reshi2022covid} to generate the continuous sentiment and subjectivity score of the content user posts, where the former reflects the emotional charge of a statement and the latter reflects the degree of objectivity. After that, we use the tweets' timestamp to calculate the time interval, and use the sentiment and subjectivity score including whether the news is positive (1) or not (0) as the covariates of our causal structure. We split the training set, validation set and test set according to the ratio of 80\%/10\%/10\% to train the proposed model. Specifically, we adopt the Adam optimizer with learning rate $1e-5$ and the epoch is set to be 500 with batch size 128. In addition, we add the dropout with 0.1 rate.

For the impact on text representation, we use the pre-trained BERT model~\cite{devlin2018bert} to extract the representation vectors of each tweet's content with 768 dimension. After that, we use PCA tools to reduce into 2 dimensions and use the new vectors as the covariates in the model. The training setting is  same with  the impact on sentiments and subjectivity experiments.

Note that all the experiments are finished on the NVIDIA GeForce RTX 2080 TI GPUs for the fair competition.

\begin{figure}[htp]
\centering
\subfigure[Impact on sentiment scores, estimated by FullyNN.]{
\begin{minipage}[t]{0.28\linewidth}
\centering
\includegraphics[height=2.5cm]{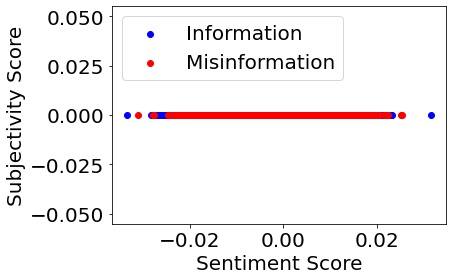}
\label{analysis_fin4}
\end{minipage}
}
\hspace{10px}
\subfigure[Impact on text representation, estimated by FullyNN.]{
\begin{minipage}[t]{0.28\linewidth}
\centering

\includegraphics[height=2.5cm]{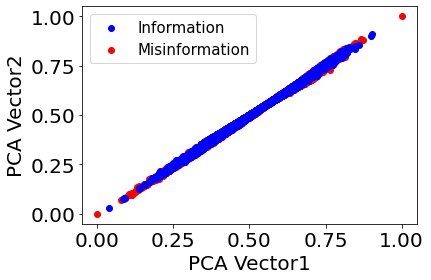}
\label{analysis_fin5}
\end{minipage}
}
\hspace{10px}
\subfigure[Visualization of raw event feature.]{
\begin{minipage}[t]{0.28\linewidth}
\centering
\includegraphics[height=2.5cm]{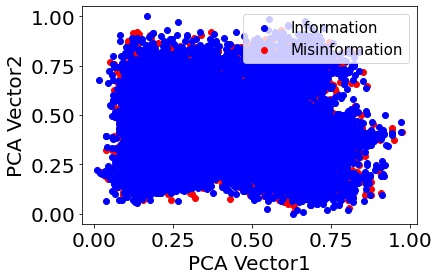}
\label{analysis_fin6}
\end{minipage}
}

\caption{Visualization of the results from baselines on real world}
\label{analysis_fin}
\end{figure}

Besides, we also evaluate FullyNN (a baseline) on the real-world dataset. Its results are visualized in Figure \ref{analysis_fin4} and \ref{analysis_fin5}. As we can see, it can not acquire recognizable differences between information and misinformation. We further visualize the raw event features (BERT embedding with dimension reduction of PCA, not finetuned on fake news detection) in Figure \ref{analysis_fin6}. As we can see, it is also hard for a pre-trained language model to distinguish information and misinformation in an unsupervised manner.

\subsection{Evaluation Metrics of Real-world Data}
\label{app:real_metric}

In real-world data experiment, we applied Normalized Average Sum of Distances. More specifically, we first normalize the scale range of each embedding dimension (ITE vector or raw input) to be same for different methods to ensure that the scales embedding spaces of different methods are comparable. Then, we calculate their Averaged Sum of Distances (ASD), which is defined as $ASD = \sum_{i=0}^k\sum_{p\in C_{i}}(p-m_{i})/n$, where $C_{i}$ is the set of ITE results (or event feature) and $i=1$ means the information results, $i=0$ means the misinforamtion results, $p$ stands for an instance in $C_i$, $m_i$ is the center of all instance representations in $C_i$, $n$ is the total number of instance.

\end{document}